\documentclass[12pt,twoside]{article}
\usepackage{graphicx,url}
\usepackage[british]{babel}
\usepackage{amsmath,amssymb}
\usepackage[tight]{subfigure}
\usepackage{sidecap}
\usepackage{algorithmic}
\usepackage{algorithm}
\def\eqvsp{}  \newdimen\paravsp  \paravsp=1.3ex
\def\,{\mskip 3mu} \def\>{\mskip 4mu plus 2mu minus 4mu} \def\;{\mskip 5mu plus 5mu} \def\!{\mskip-3mu}
\def\dispmuskip{\thinmuskip= 3mu plus 0mu minus 2mu \medmuskip=  4mu plus 2mu minus 2mu \thickmuskip=5mu plus 5mu minus 2mu}
\def\textmuskip{\thinmuskip= 0mu                    \medmuskip=  1mu plus 1mu minus 1mu \thickmuskip=2mu plus 3mu minus 1mu}
\textmuskip
\def\beq{\\\eqvsp\dispmuskip\begin{equation}}    \def\eeq{\eqvsp\end{equation}\textmuskip\\}
\def\beqn{\eqvsp\dispmuskip\begin{displaymath}}\def\eeqn{\eqvsp\end{displaymath}\textmuskip}
\def\bqa{\eqvsp\dispmuskip\begin{eqnarray}}    \def\eqa{\eqvsp\end{eqnarray}\textmuskip}
\def\bqan{\eqvsp\dispmuskip\begin{eqnarray*}}  \def\eqan{\eqvsp\end{eqnarray*}\textmuskip}

\def\paradot#1{\vspace{\paravsp plus 0.5\paravsp minus 0.5\paravsp}\noindent{\bf\boldmath{#1.}}}
\def\paranodot#1{\vspace{\paravsp plus 0.5\paravsp minus 0.5\paravsp}\noindent{\bf\boldmath{#1}}}
\def\req#1{(Equation \ref{#1})}

\def\Euro{$\,$C$\!\!\!\!\!$\raisebox{0.2ex}{=}}

\def\SetR{I\!\!R}

\def\SetZ{Z\!\!\!Z}

\def\v{\boldsymbol}

\def\X{\v\theta}
\def\Xo{\v\theta_0}
\def\GSF{\v{G_I}}
\def\GSFuv{{G_I}(u,v)}
\def\GSN{\v{G_N(\v\theta)}}
\def\GSNuv{{G_N(\v\theta)(u,v)}}

\def\F{\v I}
\def\Fuv{I(u,v)}
\def\W{\v W}
\def\Wuv{W(u,v)}
\def\Wiuv{W_i(u,v)}
\def\Nx{\Phi_x(u,v,\X)}
\def\Ny{\Phi_y(u,v,\X)}
\def\Nz{\Phi_z(u,v,\X)}

\def\Nxbyu{\frac{\partial\Nx}{\partial{u}}}
\def\Nxbyv{\frac{\partial\Nx}{\partial{v}}}
\def\Nybyu{\frac{\partial\Ny}{\partial{u}}}
\def\Nybyv{\frac{\partial\Ny}{\partial{v}}}
\def\Nzbyu{\frac{\partial\Nz}{\partial{u}}}
\def\Nzbyv{\frac{\partial\Nz}{\partial{v}}}
\def\Fbyu{\frac{\partial\F}{\partial{u}}}
\def\Fbyv{\frac{\partial\F}{\partial{v}}}
\def\Iouv{\v {I}}
\def\Io{\v {I}}
\def\Gsuv{{G_\sigma}}
\def\Gs{\v {G_\sigma}}
\def\gsmooth{Gaussian smoothing}

\begin{document}
\title{\vspace{-4ex}
\vskip 2mm\bf\Large\hrule height5pt \vskip 4mm
3D Model Assisted Image Segmentation
\vskip 4mm \hrule height2pt}
\author{{\bf Srimal Jayawardena} and {\bf Di Yang} and {\bf Marcus Hutter}\\[3mm]
\normalsize Research School of Computer Science\\[-0.5ex]
\normalsize Australian National University \\[-0.5ex]
\normalsize Canberra, ACT, 0200, Australia \\
\normalsize \texttt{\{srimal.jayawardena, di.yang, marcus.hutter\}@anu.edu.au}
}
\date{December 2011}
\maketitle

\begin{abstract}
The problem of segmenting a given image into coherent regions is important
in Computer Vision and many industrial applications require segmenting  a known object into its components.
Examples include identifying individual parts of a component for process control work in a manufacturing plant and identifying parts of a car from a photo for automatic damage detection.
Unfortunately most of an object's parts of interest in such applications share the same pixel characteristics, having similar colour and texture.
This makes segmenting the object into its components a non-trivial task for conventional image segmentation algorithms.
In this paper, we propose a ``Model Assisted Segmentation'' method to tackle this problem.
A 3D model of the object is registered over the given image by optimising a novel gradient based loss function.
This registration obtains the full 3D pose from an image of the object.
The image can have an arbitrary view of the object and is not limited to a particular set of views.
The segmentation is subsequently performed  using a level-set based method, using the projected contours of the registered 3D model  as initialisation curves.
The method is fully automatic and requires no user interaction.
Also, the system does not require any prior training.
We present our results on photographs of a real car.

\paradot{Keywords}
Image segmentation;
3D-2D Registration;
3D Model; Monocular;
Full 3D Pose;
Contour Detection;
Fully Automatic.
\end{abstract}

\section{Introduction}

\begin{figure}[t]
\centering
\def\w{0.48}
\def\testpath{T260_SANY2384_800x600w_png_contours_}
  \includegraphics[trim=5px 45px 5px 45px,, clip, width=\w\textwidth]{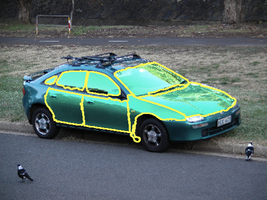}
\caption[Result and benchmark comparison 4]
{
The figure shows `Model Assisted Segmentation' results for a semi-profile view of the car.
}
\label{fig:Contours4}
\end{figure}

Image segmentation is a fundamental problem in computer vision.
Most standard image segmentation techniques rely on exploiting differences between pixel regions such as  color and texture.
Hence, segmenting sub-parts of an object which have similar characteristics can be a daunting task.
We propose a method that performs such sub-segmentation and does not require user interaction or prior training.
A result from our method is shown in Figure \ref{fig:Contours4} with the car sub-segmented into a collection of parts.
This includes the hood of the car, windshield, fender, front and back doors/windows.

Many industry applications require an image of a known object to be sub-segmented and separated into its parts.
Examples include identification of individual parts of a car given a photograph for automatic damage identification or the identification of sub-parts of a component in a manufacturing plant for process control work.
Sub-segmenting parts of an object which share the same color and texture is very hard, if not impossible, with conventional segmentation methods.
However, prior knowledge of the shape of the known object and its components can be exploited to make this task easier.
Based on this rationale we propose a novel \emph{Model Assisted Segmentation} method for image segmentation.

We propose to register a 3D model of the known object over a given photograph/image in order to initialise the segmentation process.
The segmentation is performed over each part of the object in order  to obtain sub-segments from the image.
A major contribution of this work is a novel gradient based loss function, which is used to estimate the full 3D pose of the object in the given image.
The projected parts of the 3D model may not perfectly match the corresponding parts in the photo due to dents in a damaged vehicle or inaccuracies in the 3D model.
Therefore, a level-set  \cite{li2005level} based segmentation method is initialised using initial contour information obtained by projecting parts of the 3D model at this 3D pose.
We focus our work on sub-segmentation of known car images.
Cars pose a difficult segmentation task due to highly reflective surfaces in the car body.
The method can be adapted to work for any object.

The remainder of this paper is organised as follows.
Previous work related to our paper is described in Section \ref{RelatedWork}.
We describe the method used to estimate the 3D pose of the object in Section \ref{3DRegistration}.
The contour based image segmentation approach is described next in Section \ref{CountourDetection}.
This is followed by results on real photos which are benchmarked against state of the art methods in Section \ref{Results}.

\section{Related Work}\label{RelatedWork}

Model based object recognition has received considerable
attention in computer vision. A survey by Chin and Dyer
\cite{modelbasedsurvey1986} shows that model based object
recognition  algorithms generally fall into three categories,
based on the  type of object representation used - namely 2D
representations, 2.5D representations and 3D representations.

\paranodot{2D}
representations \cite{perkins1978,yachida1977} aim to identify the
presence and orientation of a specific face of 3D objects, for
example parts on a conveyor belt. These approaches require prior
training to determine which face to match to, and are unable to
generalise to other faces of the same object.

\paranodot{2.5D}
approaches \cite{poje1982,horn1975,forsythe} are also viewer
centred, where the object is known to occur in a particular view.
They differ from the 2D approach as the model stores additional
information such as intrinsic image parameters and
surface-orientation maps.

\paranodot{3D}
approaches are utilised in situations where the object of
interest can appear in a scene from multiple viewing angles. Common
3D representation approaches can be either an `exact representation'
or a `multi-view feature representation'. The latter method uses a
composite model  consisting of  2D/2.5D models for a limited set of
views. Multi-view feature representation is used along with the
concept of generalised cylinders  by Brooks and Binford
\cite{brookes1981} to detect different types of industrial motors in
the so  called ACRONYM system. The models used in the exact
representation method, on the contrary, contain an exact
representation of the complete 3D object. Hence a 2D projection of
the object can be created for any desired  view. Unfortunately, this
method is often considered too costly in terms of processing time.
The 2D and 2.5D representations are insufficient for general
purpose applications. For example, a vehicle may be
photographed from an arbitrary view in order to indicate the
damaged parts.
Similarly, the 3D multi-view feature representation is also not
suitable, as we are not able to limit the pose of the vehicle to a
small finite set of views. Therefore, pose identification has to be done
using an exact 3D model. Little work has been done to date on
identifying the pose of an exact 3D model from a single 2D image.

\paradot{Image gradients}
Gray scale image gradients have been used to estimate the 3D pose in
traffic  video footage from a stationary camera by Kollnig and Nagel
\cite{3dposegrayval}.  The method compares image gradients  instead
of simple edge segments, for better performance. Image gradients
from projected polyhedral models are compared against image
gradients in video images.
The pose is formulated using three degrees of freedom; two for position
and one for angular orientation. Tan and Baker \cite{tan2000efficient}
use image gradients and a Hough transform based algorithm for
estimating  vehicle pose in traffic scenes, once more describing the
pose via three degrees of freedom. Pose estimation using three degrees of
freedom  is adequate for traffic image sequences, where the camera
position remains fixed with respect to the ground plane.
This approach does not recover the full 3D pose as in  our method.

\paranodot{Feature-based methods}
\cite{david2004softposit,moreno2008pose} attempt to simultaneously solve  the pose and
point correspondence problems. The success of these methods are
affected by the quality of the features extracted from the object,
which is non-trivial with objects like cars.
Features depend  on the object geometry and can cause problems when recovering a full 3D pose.
Also different image modalities cause problems with feature based methods.
For example reflections which may appear as image features do not occur in the 3D model projection.
Our method on the contrary, does not depend on feature extraction.

\paradot{Segmentation}
The use of shape priors for segmentation and pose estimation have been investigated in \cite{3DModelJointPoseSeg, compContourPoseShape, poseFreeContour, rousson2002shape}.
These methods focus on segmenting foreground from background using 3D free-form contours.
Our method, on the contrary, does \emph{intra-object} segmentation (into sub-segments) by initialising the segmentation using projections of 3D CAD model parts at an estimated pose.
In addition, our method works on more complex objects like real cars.

\section{3D Model Registration}\label{3DRegistration}

We describe the use of a featureless gradient based loss function which is used to register the 3D model over the 2D photo.
Our method works on triangulated 3D CAD models with a large number of polygons (including 3D models obtained from laser scans) and utilises image gradients of the 3D model surface normals rather than considering simple edge segments.

\paradot{Gradient based loss function}
We define a gradient based loss function that has a minimum at the correct 3D pose $\Xo \in \SetR^7$
where the projected 3D model matches the object in the given photo/image.
The image gradients of the 3D model surface normal components
and the image gradients of the 2D photo are used to define a loss function at a given pose $\X$.

We  use $(u, v) \in \SetZ^2$ to denote 2D pixel coordinates in the
photo/image and $(x, y, z) \in \SetR^3$ to denote  3D coordinates of the 3D model.
Let $\W$ be a $d$ dimensional  matrix (for example $d=3$ if $\W$ is an RGB image) with elements $W(u,v) \in \SetR^d$.
We define the $k$ norm `gradient magnitude' matrix of $\W$ as
\beq\label{eqn:gs_W}
\begin{array}{l}
  || \nabla{\Wuv} ||_k^k := \sum_{i=1}^d \left( | \frac{\partial{\Wiuv}}{\partial{u}}|^k
                   + | \frac{\partial{\Wiuv}}{\partial{v}}|^k \right)
\end{array}
\eeq
Based on this we have the gradient magnitude matrix $\GSF$ for a 2D photo/image $\F$  as
\beq\label{eqn:gs_F}
\begin{array}{l}
  \GSFuv = || \nabla{\Fuv} ||_k^k
\end{array}
\eeq
Let $\v\phi(x,y,z, \X)= \left( \phi_x \; \phi_y \; \phi_z \right)^T \in \SetR^3 $  be the unit surface normal
at the 3D point  $p = (x, y, z)$ for the 3D model at pose $\X$.
The model is rendered with the surface normal components values $\phi_x$, $\phi_y$ and $\phi_z$ used as RGB color values in the OpenGL renderer to obtain the projected surface normal component matrix $\v\Phi$ such that $\Phi(u,v, \X) \in \SetR^3$  has surface normal component values at the 2D point $(u,v)$ in the projected image.
Based on this we have the gradient normal matrix for the surface normal components as
\beq\label{eqn:gs_N}
\begin{array}{l}
  \GSNuv = || \nabla{\Phi(u,v, \X)} ||_k^k
\end{array}
\eeq
The loss function $L_g(\v\theta)$ for a given pose $\v\theta$ is defined as
\beq\label{eqn:Gradloss}
\begin{array}{l}
  L_g(\v\theta) := 1 - (corr(\GSN,\GSF))^2  \in [0,1]
\end{array}
\eeq
where $corr(\GSN,\GSF)$ is the Pearson's product-moment correlation coefficient \cite{pearsonscorrelation} between the matrix elements of $\GSN$ and $\GSF$.
This loss has a convenient property of ranging between $0$ and $1$.
Lower loss values imply a better 3D pose.

\begin{figure}[htp]
\def\tpath{}
\centering
 \def\varWidth{0.48} \subfigure[$\Nx$]{\label{fig:Mazda3SurfaceNormals0nx}
  \includegraphics[trim = 130px 200px 130px 210px, clip, width=\varWidth\textwidth]{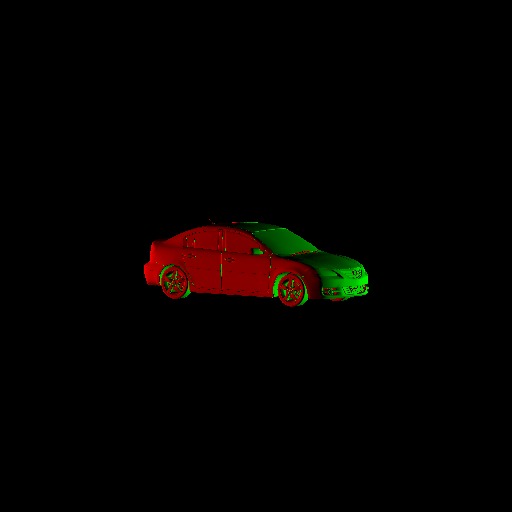}}
\subfigure[$\Ny$]{\label{fig:Mazda3SurfaceNormals0ny}
  \includegraphics[trim = 130px 200px 130px 210px, clip, width=\varWidth\textwidth]{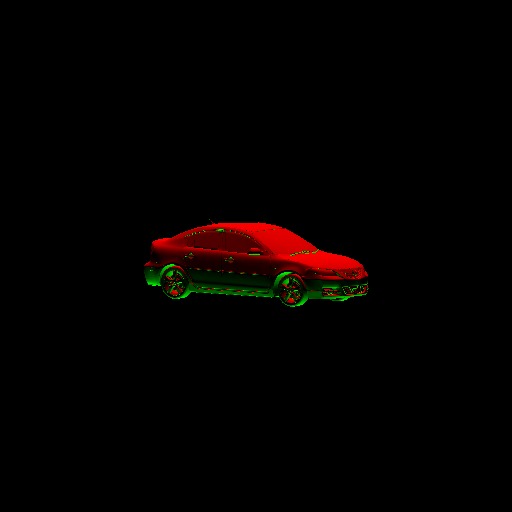}}
\subfigure[$\Nz$]{\label{fig:Mazda3SurfaceNormals0nz}
  \includegraphics[trim = 130px 200px 130px 210px, clip, width=\varWidth\textwidth]{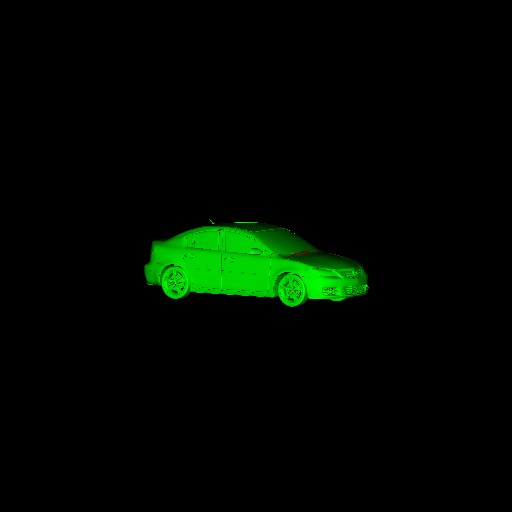}}
\subfigure[$\Nxbyu$]{\label{fig:Mazda3SurfaceNormals0nNxbyu}
  \includegraphics[trim = 130px 200px 130px 210px, clip, width=\varWidth\textwidth]{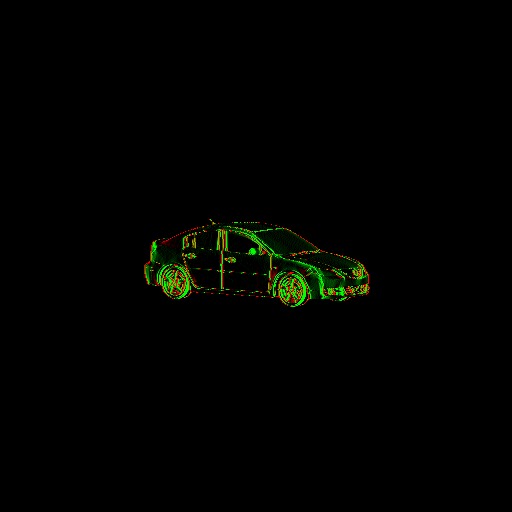}}
\subfigure[$\Nxbyv$]{\label{fig:Mazda3SurfaceNormals0nNxbyv}
  \includegraphics[trim = 130px 200px 130px 210px, clip, width=\varWidth\textwidth]{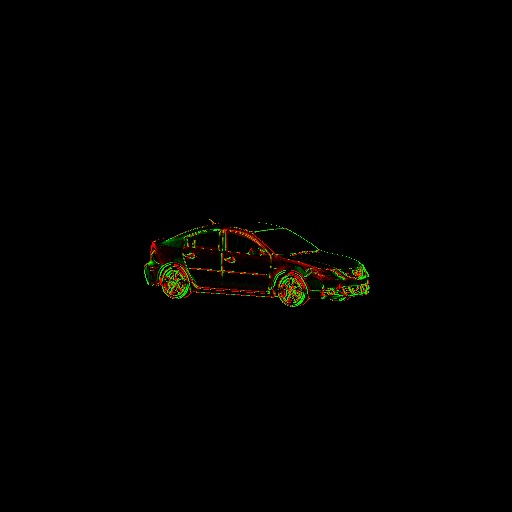}}
\subfigure[$\Nybyu$]{\label{fig:Mazda3SurfaceNormals0nNybyu}
  \includegraphics[trim = 130px 200px 130px 210px, clip, width=\varWidth\textwidth]{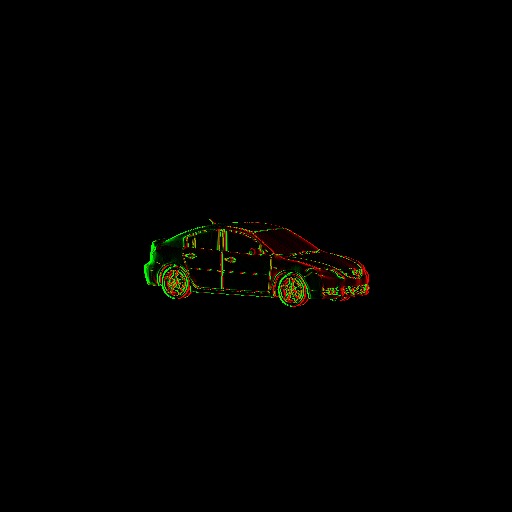}}
\subfigure[$\Nybyv$]{\label{fig:Mazda3SurfaceNormals0nNybyv}
  \includegraphics[trim = 130px 200px 130px 210px, clip, width=\varWidth\textwidth]{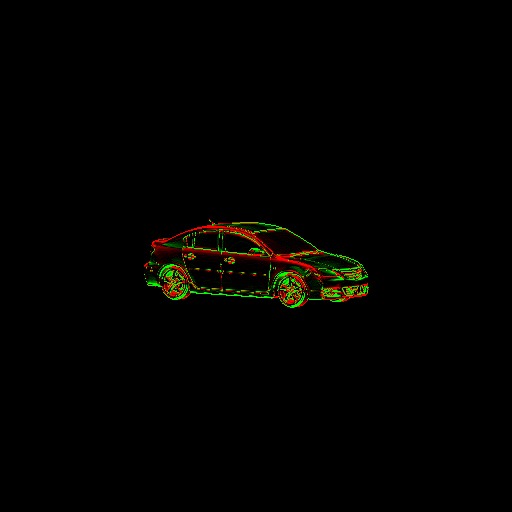}}
\subfigure[$\Nzbyu$]{\label{fig:Mazda3SurfaceNormals0nNzbyu}
  \includegraphics[trim = 130px 200px 130px 210px, clip, width=\varWidth\textwidth]{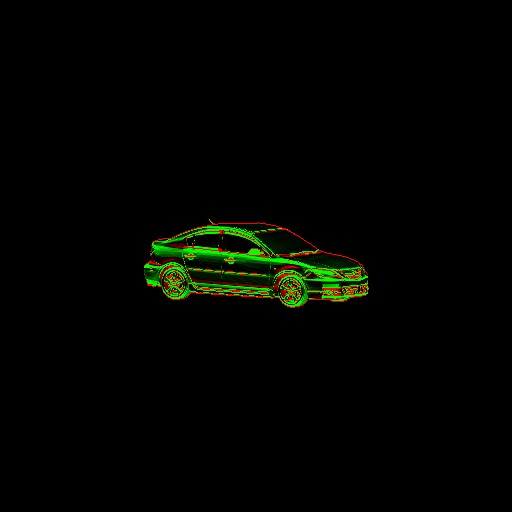}}
\subfigure[$\Nzbyv$]{\label{fig:Mazda3SurfaceNormals0nNzbyv}
  \includegraphics[trim = 130px 200px 130px 210px, clip, width=\varWidth\textwidth]{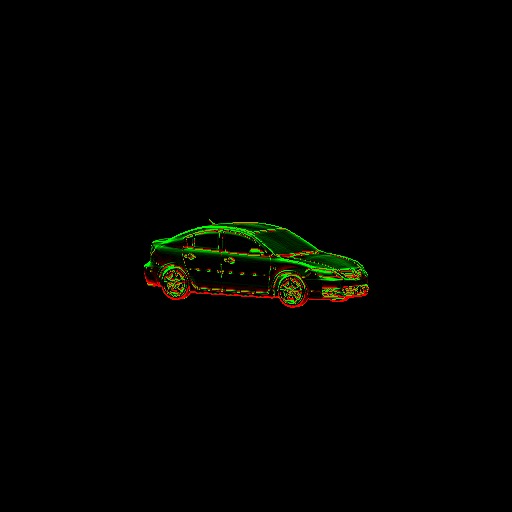}}
\subfigure[$\GSN$]{\label{fig:Mazda3SurfaceNormals0nGSN}
  \includegraphics[trim = 130px 200px 130px 210px, clip, width=\varWidth\textwidth]{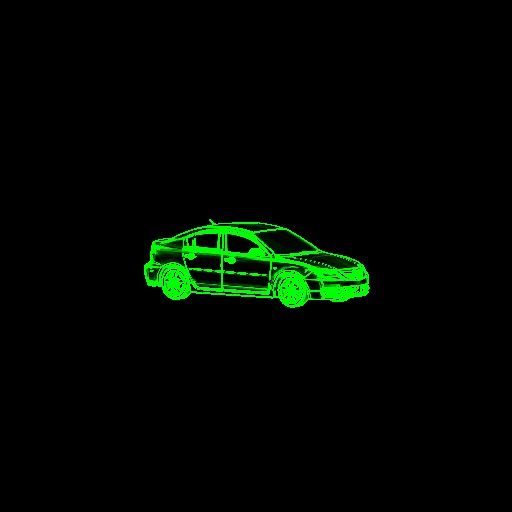}}
\caption[Visualisations of surface normal components for the Mazda3 with $n=0$]
{
The visualisations shows $\GSN$  for a 3D model in \subref{fig:Mazda3SurfaceNormals0nGSN}. The x,y and z component matrices of the surface normal vector are shown in \subref{fig:Mazda3SurfaceNormals0nx}-\subref{fig:Mazda3SurfaceNormals0nz}.

Their image gradients are shown in \subref{fig:Mazda3SurfaceNormals0nNxbyu}-\subref{fig:Mazda3SurfaceNormals0nNzbyv}.
The resulting $\GSN$ matrix is shown in \subref{fig:Mazda3SurfaceNormals0nGSN}.
No {\gsmooth} has been applied.

Colour representation: green=positive, black=zero and red=negative.
We use a horizontal $x$ axis pointing left to right, vertical $y$ axis and pointing top to bottom and an $z$ axis which points out of the page.
}
\label{fig:Mazda3SurfaceNormals0}
\end{figure}

\begin{figure}[htp]
 \def\varWidth{0.48} \def\testSpecular{specular_}
\def\testReal{real_}
\centering
\subfigure[Real photo]{\label{fig:Mazda3PhotoGradientsReal}
  \includegraphics[trim = 25px 45px 2.5px 45px, clip,width=\varWidth\textwidth]{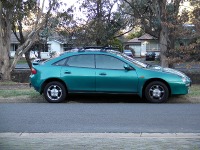}}
\subfigure[Synthetic photo]{\label{fig:Mazda3PhotoGradients0Specular}
\includegraphics[trim = 130px 200px 130px 210px, clip, width=\varWidth\textwidth]{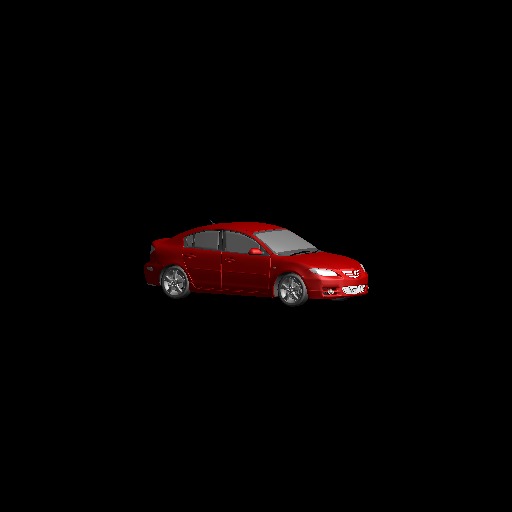}}
\subfigure[Real $\Fbyu$]{
\includegraphics[trim = 100px 180px 10px 180px, clip, width=\varWidth\textwidth]{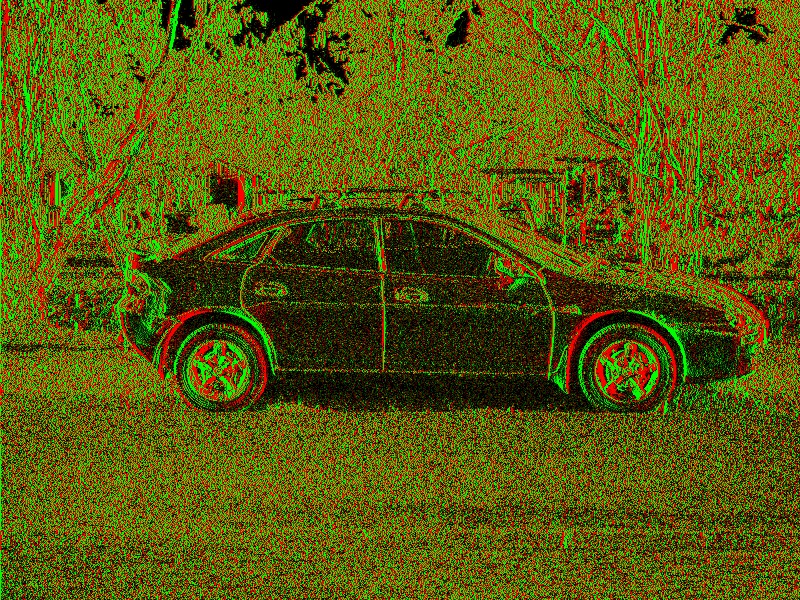}}
\subfigure[Synthetic $\Fbyu$]{
\includegraphics[trim = 130px 200px 130px 210px, clip, width=\varWidth\textwidth]{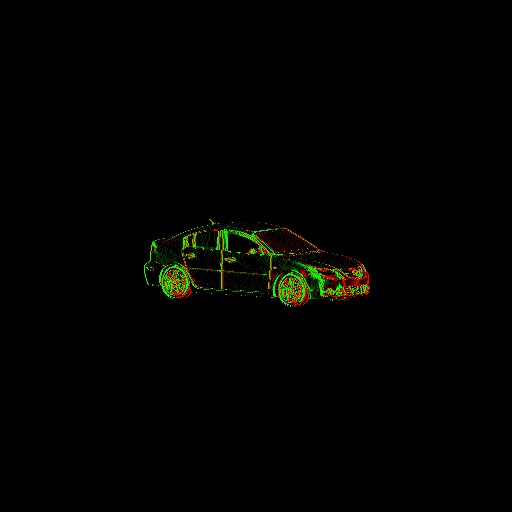}}
\subfigure[Real $\Fbyv$]{
\includegraphics[trim = 100px 180px 10px 180px, clip, width=\varWidth\textwidth]{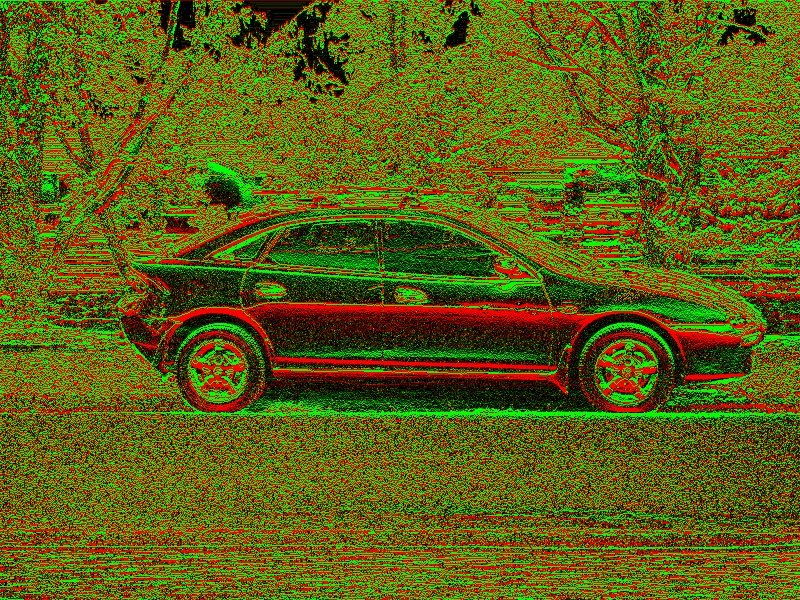}}
\subfigure[Synthetic $\Fbyv$)]{\label{fig:Mazda3PhotoGradients0FbyvSpecular}
\includegraphics[trim = 130px 200px 130px 210px, clip, width=\varWidth\textwidth]{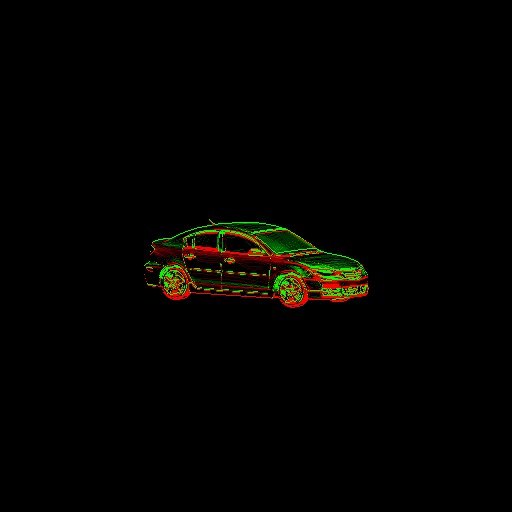}}
\subfigure[Real $\GSF$]{
\includegraphics[trim = 100px 180px 10px 180px, clip, width=\varWidth\textwidth]{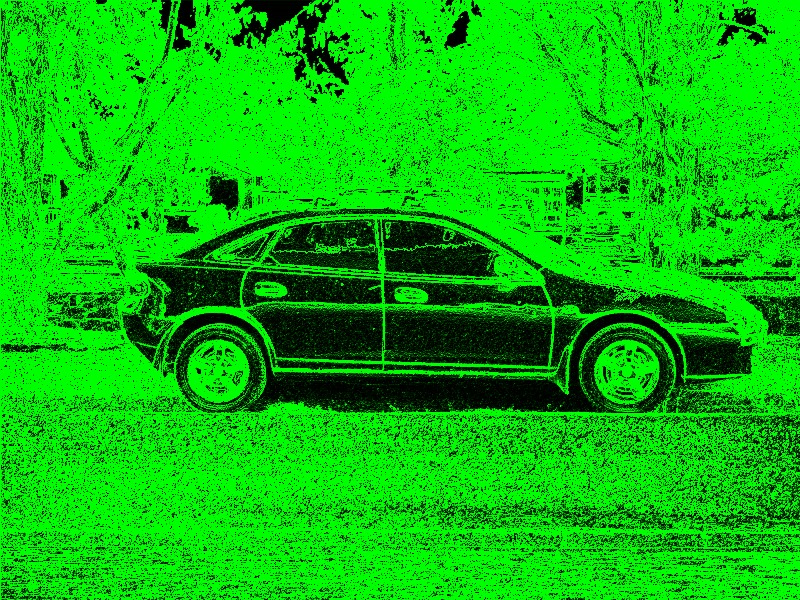}}
\subfigure[Synthetic $\GSF$]{\label{fig:Mazda3PhotoGradients0GSFSpecular}
\includegraphics[trim = 130px 200px 130px 210px, clip, width=\varWidth\textwidth]{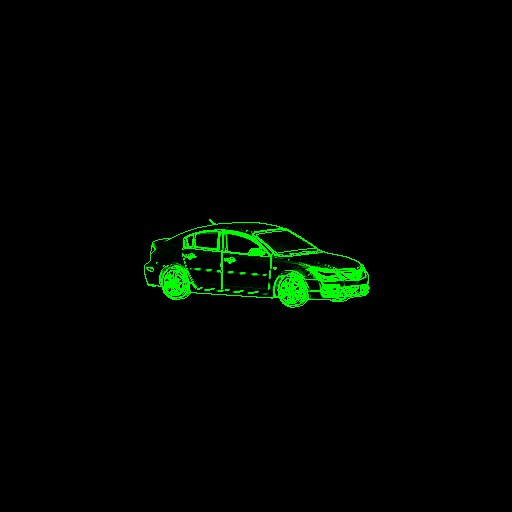}}
\caption[Visualisation of image gradients of Mazda3 photos with $n=0$]
{
Intermediate steps in calculating  $\GSF$  for a real (column 1) and synthetic photo (column 2).
The synthetic photo was made by projecting the 3D model.
Image gradients (rows 2 and 3) and $\GSF$ (row 4) are shown.
Colour representation: green=positive, black=zero and red=negative.}
\label{fig:Mazda3PhotoGradients0}
\end{figure}

\begin{figure}[htp]
\def\testSpecular{T238SpecularMazda3512x512}
\def\testReal{T260_SANY2382_800x600w_png_gradientimages2norm}
\centering
 \def\varWidth{0.3} \subfigure[Real]{
  \includegraphics[trim = 25px 45px 2.5px 45px, clip,width=\varWidth\textwidth]{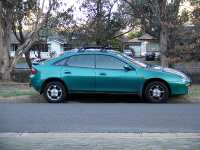}}
\subfigure[n=0]{\label{fig:Mazda3OverlayFlat0}
  \includegraphics[trim = 100px 180px 10px 180px, clip,  width=\varWidth\textwidth]{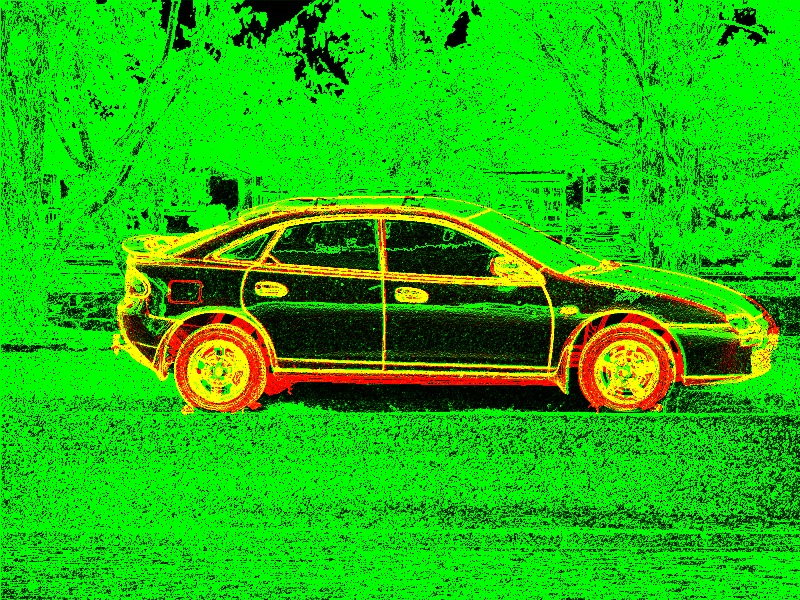}}
\subfigure[n=2]{\label{fig:Mazda3OverlayFlat3}
  \includegraphics[trim=25px 45px 2.5px 45px,, clip, width=\varWidth\textwidth]{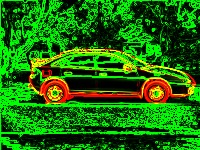}}
\subfigure[Synthetic]{
  \includegraphics[trim=130px 200px 130px 210px,, clip, width=\varWidth\textwidth]{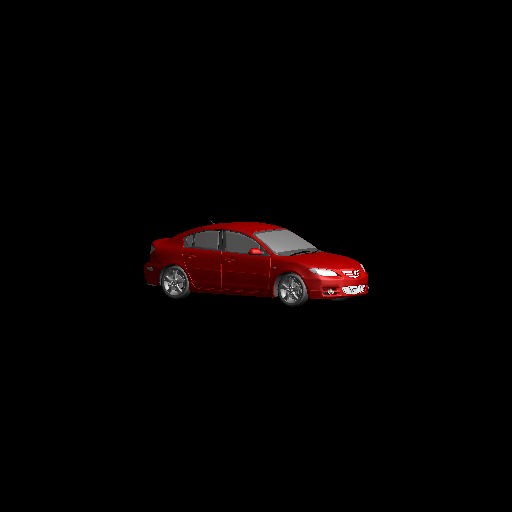}}
\subfigure[n=0]{\label{fig:Mazda3OverlaySpecular0}
  \includegraphics[trim=130px 200px 130px 210px,, clip, width=\varWidth\textwidth]{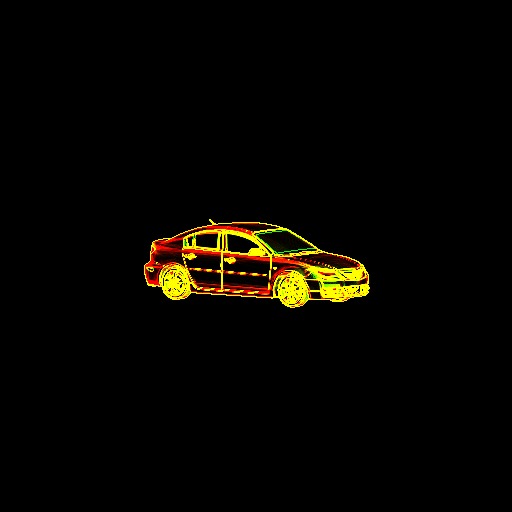}}
\subfigure[n=2]{\label{fig:Mazda3OverlaySpecular3}
  \includegraphics[trim=30px 48px 30px 52px,, clip, width=\varWidth\textwidth]{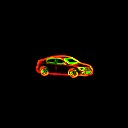}}
\caption[Overlays of $\GSN$ and $\GSN$ for Mazda3 photos .]
{Overlaid images of $\GSF$ and $\GSN$ for a real photo (row 1) and a synthetic photo (row 2) obtained by rendering a 3D model  are shown.
The first column shows the photos $\F$.
The overlaid images of $\GSF$ and $\GSN$ with no {\gsmooth} (column 2) and 3 levels of {\gsmooth} (column 3) are shown.
The photo is in the green channel and 3D model is in the red channel, with yellow showing overlapping regions.}
\label{fig:Mazda3Overlay}
\end{figure}

\paradot{Visualisation}
We illustrate  intermediate steps of the loss calculation for a 3D model of a Mazda 3 car.
The surface normal components $\Nx$ $\Ny$ and $\Nz$ are shown in Figure \ref{fig:Mazda3SurfaceNormals0}(a-c).
Their image gradients are  shown in Figure \ref{fig:Mazda3SurfaceNormals0}(d-i) and the resulting $\GSN$ matrix image is shown in Figure \ref{fig:Mazda3SurfaceNormals0}(j).
Similarly intermediate steps in the calculation of $\GSF$ are show in Figure \ref{fig:Mazda3PhotoGradients0} for a real photo and a  synthetic photo.
We show overlaid images of $\GSN$ and $\GSF$ at the known matching pose $\X$ in Figure \ref{fig:Mazda3Overlay}.
We show how the overlap changes by applying $2$ levels of {\gsmooth}  (described below) in Figures \ref{fig:Mazda3Overlay} for the real and synthetic photo.
The synthetic photos were made by projecting the 3D model at a known pose $\X$.

\begin{figure}[htp]
\centering
\def\w{0.48} \subfigure[2-norm]{
\includegraphics[trim=14px 1px 10px 10px,, clip, width=\w\textwidth]{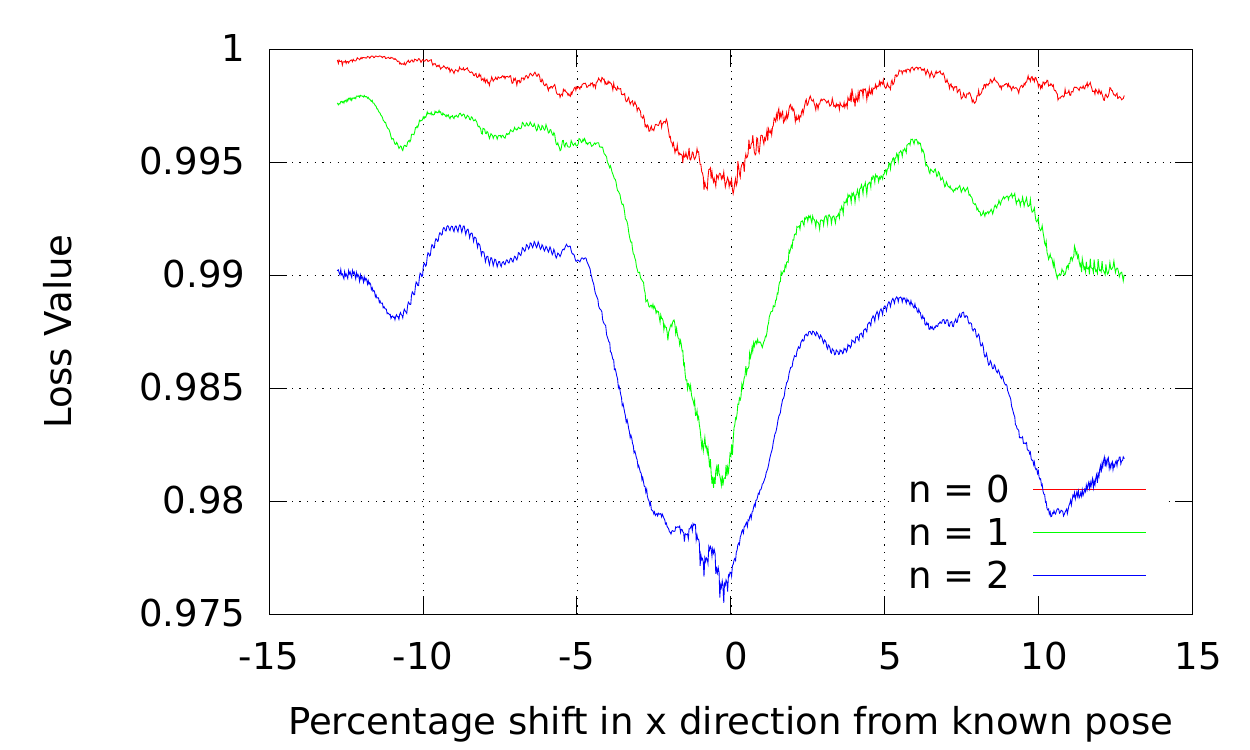}}
\subfigure[1-norm]{
\includegraphics[trim=14px 1px 10px 10px,, clip, width=\w\textwidth]{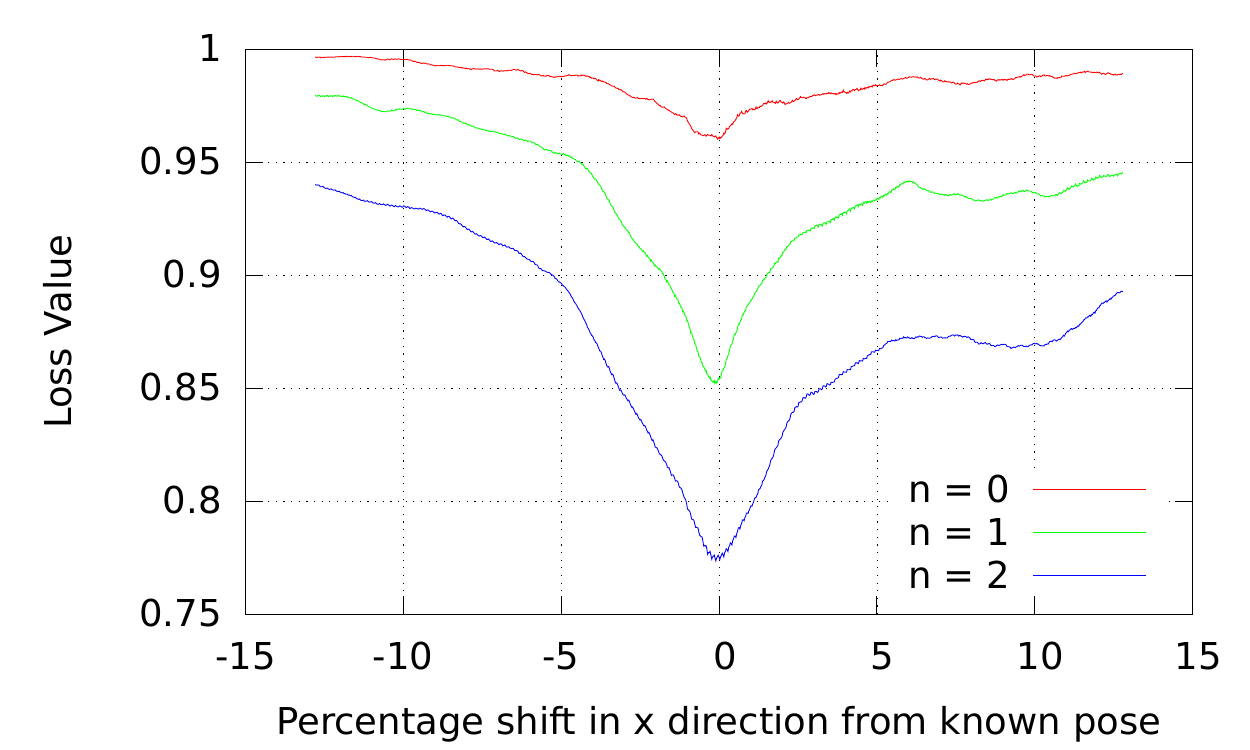}}
\caption[1norm and 2norm gradloss 1D landscapes]
{We compare $1$-norm and $2$-norm loss landscapes obtained by shifting the 3D model along the x direction from a known 3D pose.
The horizontal axis shows the percentage deviation along the x axis.
The numbers in the legend show the level of {\gsmooth} $n$ applied on the gradient images before calculating the loss in Equation \ref{eqn:Gradloss}.
We note that the $1$-norm loss is less noisy compared to the $2$-norm loss.
The actual loss function is seven dimensional and graphs of the other dimensions are similar.}
\label{fig:1norm2normExample}
\end{figure}

The correlation will be highest in Equation \ref{eqn:Gradloss} when the 3D model is projected with pose
parameters $\X_0$ that match the object in the photo $F$, as this has the best overlap.
Therefore the loss will be lowest at the correct pose parameters $\X_0$, for values of $\X$ reasonably close to $\X_0$.
We see this in the loss landscapes in Figure \ref{fig:1norm2normExample}.

\paradot{\gsmooth}
We do {\gsmooth} on the photo and rendered surface normal component images before calculating $\GSF$ (Equation \ref{eqn:gs_F}) and $\GSN$ (Equation \ref{eqn:gs_N}).
This is done by convolving with a 2D Gaussian kernel followed by down-sampling \cite{forsythe}.
This  makes the loss function landscape less steep and noisy, thus making it easier to optimise.
However, the global optimum tends to deviate slightly from the correct pose at high levels of {\gsmooth}.
Compare the 1D loss landscapes shown in Figure \ref{fig:1norm2normExample} for different levels of {\gsmooth} $n$.
Therefore, we do a series of optimisations starting from the highest level of smoothing, using the optimum found at level $n$ as the initialisation for level $n-1$, recursively.

\paradot{Choosing the norm $k$}
We have a choice when selecting the norm for Equations \ref{eqn:gs_F} and \ref{eqn:gs_N}.
Having tested both $1$-norm and $2$-norm cases  we have found the $1$-norm to be less noisy (as shown in Figure \ref{fig:1norm2normExample}) and hence easier to optimise.

\begin{figure}
\centering
\includegraphics[width=0.3\textwidth]{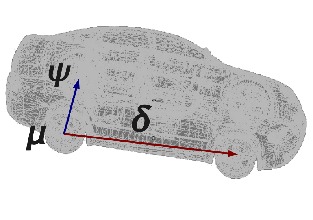}
\caption[Pose representation $\X$ used for 3D car models]{
We illustrate components of the pose representation $\X'$ (Equation \ref{eqn:PoseXRough}) used for 3D models of cars.
We use the rear wheel center $\v\mu$,  the vector between the wheel centres $\v\delta$ and unit vector $\v\psi$ in the direction of the rear wheel axle.
}
\label{fig:poseRepresentationX}
\end{figure}

\paradot{Initialisation}
We use a rough pose estimate to seed the optimisation.
An object specific method can be used to obtain the rough pose.
Possible methods for obtaining a coarse initial pose include the work done by \cite{OzuysalLF09}, \cite{depthEncodedHoughVotingEccv2010} and \cite{implicitshapepose}.
We have used the wheel match method developed by Hutter and Brewer
\cite{hutter2009matching} to obtain an initial pose for vehicle photos where the wheels are visible.
The wheels need not be visible with the other methods mentioned above.
We use the following to represent the rough pose of cars as prescribed in \cite{hutter2009matching} which neglects the effects of perspective projection.
\beq
\label{eqn:PoseXRough}
\X' := \left( \mu_x, \mu_y, \delta_x, \delta_y, \psi_x, \psi_y \right)
\eeq
$\v\mu=(\mu_x, \mu_y)$ is the visible rear wheel center of the
car in the 2D image. $\v\delta=(\delta_x, \delta_y)$ is  the
vector between corresponding rear and front wheel centres of
the car in the 2D image. The 2D image is a projection of the 3D
model on to the XY plane. $\v\psi = \left( \psi_x, \psi_y,
\psi_z \right)$ is a unit vector in the direction of the rear
wheel axle of the 3D car model. Therefore, $\psi_z = -\sqrt{1 -
\smash{\psi_x^2 - \psi_y^2}}$ and need not be explicitly
included in the pose representation $\X$. This representation
is illustrated in Figure \ref{fig:poseRepresentationX}.

We include an additional perspective parameter $f$ (the distance to the camera from the projection plane in the OpenGL 3D frustum) when optimising the loss function to obtain the fine 3D pose.
Hence we define the full 3D pose as follows.
\beq
\label{eqn:PoseXFine}
\X := \left( \mu_x, \mu_y, \delta_x, \delta_y, \psi_x, \psi_y, f \right)
\eeq
$\X'$ is converted to translation, scale and rotation as per \cite{hutter2009matching} to transform the 3D model and along with $f$ is used to render the 3D model with perspective projection in OpenGL using pose $\X$.
Thereby, we estimate the full 3D pose by minimizing Equation \ref{eqn:Gradloss} w.r.t $\X$. Intrinsic camera parameters need not be known explicitly.
Note that any other choice of pose parameters would do.
We use the above as it is convenient with cars.

\paradot{Background removal}
As the effects of the background clutter in the photo adds considerable noise to the loss function landscape we use an adaptation of the Grabcut \cite{grabcut} method to remove a considerable amount of the background pixels from the photo.
Although, this does not result in a perfect removal of the background it significantly improves the pose estimation results.
The initial rough pose estimate is used as a prior to generate the background and foreground grabcut masks \footnote{We use the cv::grabCut() method provided in OpenCV\cite{opencv_library} version 2.1}.
Figure \ref{resultSANY2382_800x600w_grabcut} shows results of the background removal.

\paradot{Optimisation}
We use the downhill simplex optimiser \cite{nelder1965simplex} to find the pose parameters $\v\theta_0$ which give the lowest loss value for Equation \ref{eqn:Gradloss}.
This optimiser is very robust and is capable of moving out of local optima by reinitialising the simplex.
Downhill simplex does not require gradient calculations.
Gradient based optimisers would be problematic given the loss landscapes in Figure \ref{fig:1norm2normExample}.
We use the fine pose obtained thus to register the 3D model on the 2D photo.
This is used to initialise contour detection based image segmentation.

\section{Contour Detection}\label{CountourDetection}

In this section, we discuss the procedure of contour detection used to segment the known object in the image.
We use a  variation of the level set method which does not require re-initialisation \cite{li2005level} to find  boundaries of relevant object parts.

Most active contour models implement an edge-function to find boundaries.
The edge-function is a gradient dependant positive decreasing function.
A common formulation is as follows
\beq\label{eqn:contour1}
g(|\nabla \Iouv |)=\frac{1}{1+|\nabla \Gsuv \otimes \Iouv|^p}, \quad p\geq1,
\eeq

where  $\Gs \otimes \Io $ denotes a smoother version of 2D image  $\Io$,   $\Gs$  is an isotropic Gaussian kernel with standard deviation $\sigma$, and $\otimes$ is the convolution operator.
Therefore $g(|\nabla \Io|)$ will be $0$, as $\nabla \Io$  approaches infinity, \emph{i.e.}
\beq\label{eqn:contour2}
  \lim_{|\nabla \Io|\to\infty}g(|\nabla \Io|)=0, \;\mathrm{when}\,\sigma = 0.
\eeq
As per \cite{li2005level}, a Lipschitz function $\phi$ is used to represent the curve $C=\{(u,\;v)|\phi_0(u,\;v)=0\}$ such that ,
\beq\label{eqn:contour3}
  \phi_0(u,\;v)=\left\{\begin{array}{ll}
        -\rho,&\quad(u,\;v)\;\mathrm{inside\;\,contour}\;\,C\\
        \quad\!0, &\quad(u,\;v)\;\mathrm{on\;\,contour}\;\, C\\
        \quad\!\rho, &\quad(u,\;v)\;\mathrm{outside\;\,contour} \;\,C
  \end{array}\right.
\eeq
As with other level set formulations like \cite{caselles1993geometric} and \cite{malladi2002shape}, the curve $C$ is evolved using the mean curvature $\mathrm{div}\left(\nabla\phi/|\nabla\phi|\right)$ in the normal direction $|\nabla\phi|$.
Therefore the curve evolution is represented by $\partial\phi/\partial t$ as
\beq\label{eqn:contour4}
  \begin{array}{ll}

    \frac{\partial\phi}{\partial t}=|\nabla\phi|\left(\mathrm{div}\left(g(|\nabla \Io|)
    \frac{\nabla\phi}{|\nabla\phi|}\right)+\nu g(|\nabla \Io|)\right),\\
    \phi(0,\;u,\;v)=\phi_0(u,\;v)
  \in [0,\;\infty)\times\mathbb{R}^2

  \end{array}
\eeq
where the evolution of the curve is given by the zero-level curve at time $t$ of the function $\phi(t,\;x,\;y)$.
$\nu$ is a constant to ensure that the curve  evolves in the normal direction, even if the mean curvature is zero.

Theoretically, as the image gradient on an edge/boundary of an image segment tends to infinity,  the edge function $g$ \req{eqn:contour1} is zero on the boundary.
This causes the curve $C$ to stop evolving at the boundary \req{eqn:contour4}.
However, in practice the edge function may not always be zero at image boundaries of complex images  and the performance of the level set method is severely affected by noise.
Isotropic Gaussian smoothing can be applied to reduce image noise but over smoothing will also smooth the edges, in which case, the level set curve may miss the boundary altogether.
This is a common problem not only for the level set method in \cite{li2005level} but also for other \emph{active contour models} \cite{caselles1993geometric,malladi1993topology,malladi1994evolutionary,malladi2002shape}.
Additionally, the efficiency and effectiveness of level set in boundary detection depends a lot on the initialisation of the curve.
Without appropriate initialisation, the curve is frequently trapped into local minima.

A very close  initialisation curve can eliminate this problem.
In our approach, the initialisation curve is obtained by registering a 3D model over the photo as described in Section \ref{3DRegistration}.
Since the parts $p$ in the 3D model are already known, they can be projected at the known 3D pose $\X$ to obtain a selected part outline $o_p$ in 2D.
An `erosion' morphological operator is applied on $o_p$ to obtain the initial curve $\phi_{0,p}$ which is inside the real boundary.

The green curves (initialisation images in Figures \ref{fig:Contours1}, \ref{fig:Contours2n3} and \ref{fig:Contours4n5n6})  are used to denote the 2D outlines of projected parts in the 3D model, while the red curves are the initialisation curves obtained by eroding these green curves.
The level set starts with the initial curve $\phi_{0,p}$ to find actual boundary $\phi_{r,p}$ in the 2D image of vehicle, for each part $p$.
The yellow curves (result images in Figures \ref{fig:Contours1}, \ref{fig:Contours2n3} and \ref{fig:Contours4n5n6})
indicate the actual boundaries  detected.

The entire process of `Model Assisted Segmentation' is given in pseudo-code in Algorithm \ref{alg:pseudocode}.
\begin{algorithm}                      \caption{Model Assisted Segmentation}          \label{alg:pseudocode}
\renewcommand{\algorithmicrequire}{\textbf{Input:}}
\renewcommand{\algorithmicensure}{\textbf{Output:}}
\begin{algorithmic}[1]
\REQUIRE Let $\F =$ Given image, $\v M =$ Known 3D model
\ENSURE Segmentation curves $\phi_{r,p}$ for selected model parts $p$
\STATE $\X' \gets$   Rough pose from $\F$
\STATE $\F' \gets$  Remove background in $\F$ using $\X'$
\STATE $\X \gets \X'$
\FOR{$n = 2$ down to 0}
\STATE $\X \gets $ Optimise $ L_g(\v\theta)$ on $\F'$ starting from $\X$ using $n$ levels of {\gsmooth}
\ENDFOR
\FOR{$p \in $ Selected parts in $\v M$}
\STATE $o_p \gets $ Outline of $p$ projected using $\X$
\STATE $\phi_{0,p} \gets $ Apply erosion operation on $o_p$
\STATE $\phi_{r,p} \gets $ Output of level set on $\F$ using $\phi_{0,p}$ as initial curve
\ENDFOR
\end{algorithmic}
\end{algorithm}

\section{Results} \label{Results}

\begin{figure}[htp]
\centering
\def\w{0.48} \def\wz{0.3}\def\testpath{T260_SANY2382_800x600w_png_}
\subfigure[Photo]{\label{resultSANY2382_800x600w_photo}
\includegraphics[width=\wz\textwidth]
{SANY2382_800x600small}}
\subfigure[Background removed]{\label{resultSANY2382_800x600w_grabcut}
  \includegraphics[trim=50px 105px 5px 90px, clip, width=\w\textwidth]{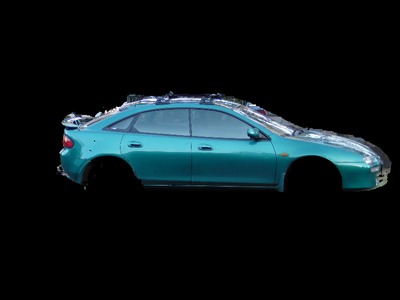}}
\subfigure[Rough pose]{\label{resultSANY2382_800x600w_rough}
  \includegraphics[trim=50px 105px 5px 90px, clip, width=\w\textwidth]{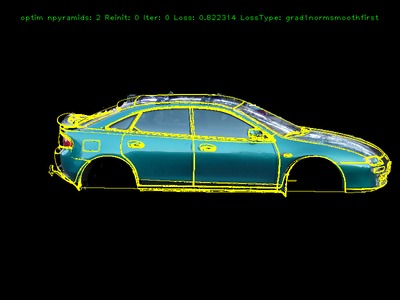}}
\subfigure[Fine pose n=2]{\label{resultSANY2382_800x600w_pyr2}
  \includegraphics[trim=50px 105px 5px 90px, clip, width=\w\textwidth]{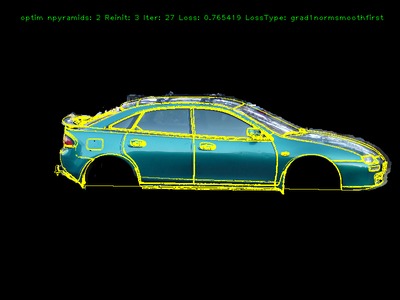}}
\subfigure[Fine pose n=1]{\label{resultSANY2382_800x600w_pyr1}
  \includegraphics[trim=50px 105px 5px 90px, clip, width=\w\textwidth]{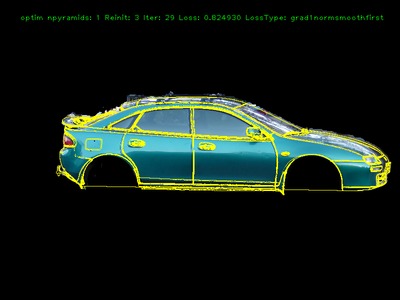}}
\subfigure[Final fine pose n=0]{\label{resultSANY2382_800x600w_pyr0}
  \includegraphics[trim=50px 105px 5px 90px, clip, width=\w\textwidth]{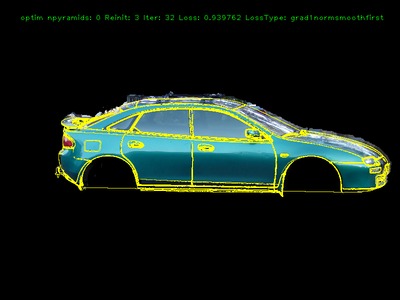}}
\caption[Pose estimation results]
{
The images show pose estimation results for a real photograph of a Mazda Astina car.
The original photograph and subsequent images have been cropped for clarity.
The fine 3D pose in \subref{resultSANY2382_800x600w_pyr0} is obtained by optimising the novel gradient based loss function (Equation \ref{eqn:Gradloss}) using the rough pose in \subref{resultSANY2382_800x600w_rough}.
The rough pose is obtained  as prescribed in \cite{hutter2009matching}.
Much of the background is removed \subref{resultSANY2382_800x600w_grabcut} from the original photo \subref{resultSANY2382_800x600w_photo} using an adaptation of `Grabcut' \cite{grabcut} when estimating the fine 3D pose.
Intermediate steps of optimising the loss function with different levels of {\gsmooth} $n$ applied on the gradient images are shown in \subref{resultSANY2382_800x600w_pyr2}, \subref{resultSANY2382_800x600w_pyr1} and \subref{resultSANY2382_800x600w_pyr0}.
The close ups highlight the visual improvement during intermediate steps Figure \ref{fig:resultSANY2382_800x600wCloseups}.
\label{fig:resultSANY2382_800x600w}
}
\end{figure}
\begin{figure}[htp]
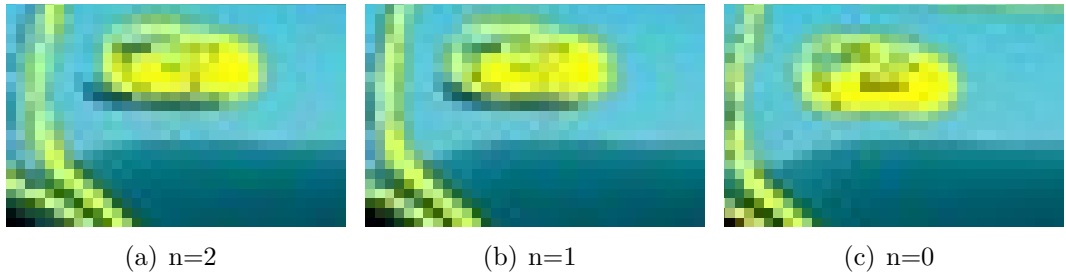

\centering
\def\wz{0.3} \def\testpath{T260_SANY2382_800x600w_png_}
\subfigure[n=2]{\label{resultSANY2382_800x600w_pyr2_closeup}
  \includegraphics[trim=120px 140px 245px 137px, clip, width=\wz\textwidth]{\testpath optim_0_3_027}}
\subfigure[n=1]{\label{resultSANY2382_800x600w_pyr1_closeup}
  \includegraphics[trim=120px 140px 245px 137px,  clip, width=\wz\textwidth]{\testpath optim_1_3_029}}
\subfigure[n=0]{\label{resultSANY2382_800x600w_pyr0_closeup}
  \includegraphics[trim=120px 140px 245px 137px,  clip, width=\wz\textwidth]{\testpath optim_2_3_032}}
\caption[Pose estimation results - close ups]
{
Close ups at each step of the optimisation (shown in Figure \ref{fig:resultSANY2382_800x600w}) for different levels of {\gsmooth} $n$ highlight the visual improvement in the 3D pose.
\label{fig:resultSANY2382_800x600wCloseups}
}
\end{figure}

\begin{figure}[htp]
\centering
\def\w{0.48} \def\testpath{T260_SANY2382_800x600w_png_contours_}
\subfigure[Initialisation]{\label{resultSANY2382_800x600w_contours_whole_init}
  \includegraphics[trim=50px 40px 30px 40px, clip, width=\w\textwidth]{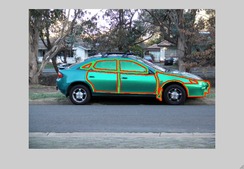}}
\subfigure[Result]{\label{resultSANY2382_800x600w_contours_whole_result}
  \includegraphics[trim=50px 40px 30px 40px, clip, width=\w\textwidth]{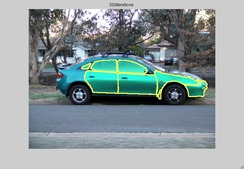}}
\caption[Result and benchmark comparison 2-3]
{
The figure shows the `Model Assisted Segmentation' results for a real photo of a Mazda Astina car.
The initialisation curves for a selection of car body parts are shown in \ref{resultSANY2382_800x600w_contours_whole_init} based on the fine 3D pose shown in Figure \ref{resultSANY2382_800x600w_pyr0}.
The 3D model outlines are shown in `green' and the initialisation curves obtained by eroding these outlines are shown in `red'.
The resulting segmentation is shown in \ref{resultSANY2382_800x600w_contours_whole_result}.
Close ups are shown along with benchmark results in Figure \ref{fig:Contours2n3}.}
\label{fig:Contours1}
\end{figure}
\begin{figure*}[t]
\centering
\def\w{0.23}
\def\testpath{T260_SANY2382_800x600w_png_contours_}
\subfigure[Initialisation]{\label{resultSANY2382_800x600w_contours_backdoor_init}
  \includegraphics[width=\w\textwidth]{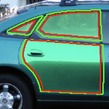}}
\subfigure[Result]{\label{resultSANY2382_800x600w_backdoor_result}
  \includegraphics[width=\w\textwidth]{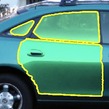}}
\subfigure[Benchmark - GC]{\label{resultSANY2382_800x600w_backdoor_benchmark}
  \includegraphics[width=\w\textwidth]{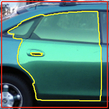}}
\subfigure[Benchmark - LS]{\label{resultSANY2382_800x600w_backdoor_benchmark_ls}
  \includegraphics[width=\w\textwidth]{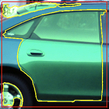}}
\subfigure[Initialisation]{\label{resultSANY2382_800x600w_contours_fender_init}
  \includegraphics[width=\w\textwidth]{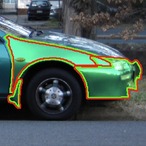}}
\subfigure[Result]{\label{resultSANY2382_800x600w_fender_result}
  \includegraphics[width=\w\textwidth]{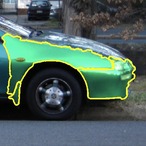}}
\subfigure[Benchmark - GC]{\label{resultSANY2382_800x600w_fender_benchmark}
  \includegraphics[width=\w\textwidth]{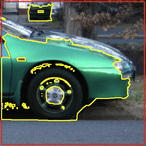}}
\subfigure[Benchmark - LS]{\label{resultSANY2382_800x600w_fender_benchmark_ls}
  \includegraphics[width=\w\textwidth]{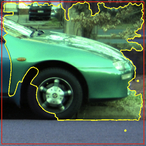}}
\caption[Result and benchmark comparison 2-3]
{
Different close ups (row wise) for the results in Figure \ref{fig:Contours1} are shown with the
initialisation curves (column 1), our results (column 2) and benchmark results (columns 3 and 4).
Our results are more accurate in general.
Note the bleeding and false positives in the benchmark results.
Our method is more accurate and sub-segments the image into meaningful parts.
}
\label{fig:Contours2n3}
\end{figure*}
\begin{figure*}[t]
\centering
\def\w{0.23}
\def\testpath{T260_SANY2384_800x600w_png_contours_}
\subfigure[Initialisation]{\label{resultSANY2384_800x600w_contours_ini_backdoor_view2}
  \includegraphics[width=\w\textwidth]{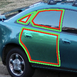}}
\subfigure[Result]{\label{resultSANY2384_800x600w_contours_res_backdoor_view2}
  \includegraphics[width=\w\textwidth]{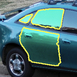}}
\subfigure[Benchmark - LS]{\label{resultSANY2384_800x600w_contours_benchmark_backdoor_view2_ls}
  \includegraphics[width=\w\textwidth]{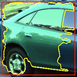}}
\subfigure[Benchmark - GC]{\label{resultSANY2384_800x600w_contours_benchmark_backdoor_view2_gc}
  \includegraphics[width=\w\textwidth]{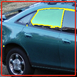}}
\subfigure[Initialisation]{\label{resultSANY2384_800x600w_contours_ini_frontdoor_view2}
  \includegraphics[width=\w\textwidth]{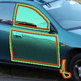}}
\subfigure[Result]{\label{resultSANY2384_800x600w_contours_res_frontdoor_view2}
  \includegraphics[width=\w\textwidth]{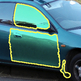}}
\subfigure[Benchmark - GC]{\label{resultSANY2384_800x600w_contours_benchmark_frontdoor_view2_gc}
  \includegraphics[width=\w\textwidth]{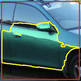}}
\subfigure[Benchmark - LS]{\label{resultSANY2384_800x600w_contours_benchmark_frontdoor_view2_ls}
  \includegraphics[width=\w\textwidth]{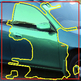}}
  \subfigure[Initialisation]{\label{resultSANY2384_800x600w_contours_ini_front_view2}
  \includegraphics[width=\w\textwidth]{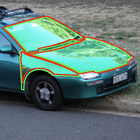}}
\subfigure[Result]{\label{resultSANY2384_800x600w_contours_res_front_view2}
  \includegraphics[width=\w\textwidth]{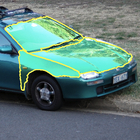}}
\subfigure[Benchmark - GC]{\label{resultSANY2384_800x600w_contours_benchmark_front_view2_gc}
  \includegraphics[width=\w\textwidth]{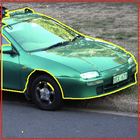}}
\subfigure[Benchmark - LS]{\label{resultSANY2384_800x600w_contours_benchmark_front_view2_ls}
  \includegraphics[width=\w\textwidth]{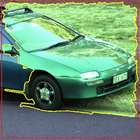}}
\caption[Results for a semi-profile view]
{
The figures show different close ups (row wise) for the results in Figure \ref{fig:Contours4}.
Initialisation curves (column 1), our results (column 2) and benchmark results (columns 3 and 4) are shown.
We note that our results more accurate and has sub-segmented the car into meaningful components.
}
\label{fig:Contours4n5n6}
\end{figure*}
We apply our method to segment components of a real car from a photograph as follows.

\paradot{Pose estimation}
The results of registering the 3D model over the photograph (pose estimation) are shown in Figure \ref{fig:resultSANY2382_800x600w}.
A gradient sketch of the 3D model is drawn over the photograph in yellow to indicate the pose of the 3D model at each step in Figure  \ref{fig:resultSANY2382_800x600w}.
The wheels of the 3D model do not match the wheels in the photo due to the effects of wheel suspension.
Since we are interested in segmenting parts of the car body the wheels have been removed from the 3D model for the fine pose estimation.
The original photograph in Figure \ref{resultSANY2382_800x600w_photo} shows the side view of a Mazda Astina car.
We register a triangulated 3D model of the car obtained by a 3D laser scan.
The rough 3D pose obtained using the wheel locations \cite{hutter2009matching} is shown in Figure  \ref{resultSANY2382_800x600w_rough}.
The result of the approximate background removal is shown in Figure \ref{resultSANY2382_800x600w_grabcut}.
We optimise the gradient based loss function (Equation \ref{eqn:Gradloss}) for the image in Figure  \ref{resultSANY2382_800x600w_grabcut} with respect to the seven pose parameters (Section \ref{3DRegistration}) to obtain the fine 3D pose.
The optimisation is done sequentially moving from the highest level of {\gsmooth} to the lowest.
We start from the rough pose with two levels of {\gsmooth} and obtain the pose in Figure \ref{resultSANY2382_800x600w_pyr2}.
Next we use this pose to initialise an optimisation of the loss function with one level of {\gsmooth} and obtain the pose in \ref{resultSANY2382_800x600w_pyr1}.
Finally, we use this pose to perform one more optimisation with no {\gsmooth} and obtain the final fine 3D pose shown in Figure  \ref{resultSANY2382_800x600w_pyr0}.
We note that the visual improvement in the image overlays gets smaller as we go up the Gaussian pyramid.
However, the improvement in the 3D pose becomes more apparent when we compare the close ups in  Figures \ref{resultSANY2382_800x600w_pyr2_closeup}, \ref{resultSANY2382_800x600w_pyr1_closeup} and   \ref{resultSANY2382_800x600w_pyr0_closeup}.

\paradot{Segmentation}
Segmentation results based on contour detection for the photograph in \ref{resultSANY2382_800x600w_photo} using the fine 3D pose (Figure \ref{resultSANY2382_800x600w_pyr0}) are shown in Figures \ref{fig:Contours1} and \ref{fig:Contours2n3}.
The segmentation results for a selection of car parts (front and back doors, front and back windows, fender, mud guard and front buffer) are shown in Figure \ref{resultSANY2382_800x600w_contours_whole_result} by the yellow curves.
The part boundaries obtained by projecting the 3D model are shown in green  and the initialisation curves are shown in red in Figure \ref{resultSANY2382_800x600w_contours_whole_init}.
For the sake of clarity we also include close ups of a few parts.
The initialisation curves and the segmentation results for the back door and window are shown in Figures \ref{resultSANY2382_800x600w_contours_backdoor_init} and \ref{resultSANY2382_800x600w_backdoor_result}, using the same color code.
Close ups for the front parts are shown in Figures \ref{resultSANY2382_800x600w_contours_fender_init} and \ref{resultSANY2382_800x600w_fender_result}.
We see the high amount of reflection in the car body deteriorating the performance of the segmentation results in the latter case, especially around the hood of the car and windshield.
In contrast the mud guard, lower parts of the buffer and fender are segmented out quite well in Figure \ref{resultSANY2382_800x600w_fender_result} as there is less reflection noise in that region.
Results for a semi-profile view of the car are shown in Figures \ref{fig:Contours4} and \ref{fig:Contours4n5n6}  using same convention.

\paradot{Accuracy}
The accuracy of the results have been compared against a ground truth obtained from the photos by hand annotation in Table \ref{tbl:accuracy}.
We calculate the accuracy as
\beq\label{eqn:accuracy}
\begin{array}{l}
a = 1 - \frac{\sum_{u,\;v}|U_R(u,\;v) - U_G(u,\;v)|}{\sum_{u,\;v}U_G(u,\;v)}
\end{array}
\eeq
where $U_R$ and $U_G$ are two binary images of the sub-segmentation result and ground truth respectively.
We note that the accuracy is considerably high.
Also, the side view has a higher accuracy in general because the pose estimation gave a better result and hence the segmentation was better initialised.

\begin{table}
\begin{center}
\begin{tabular}{|l|c|c|c|}
\hline
Part & Side View & Semi Profile & Avg.\\
\hline\hline
Fender                & 97.7\% & 97.6\% & 97.7\% \\
Front door  & 98.1\% & 95.3\% & 96.7\% \\
Back door   & 96.8\% & 93.6\% & 95.2\% \\
Mud flap      & 97.3\% & 95.1\% & 96.2\% \\
Front window& 97.8\% & 97.5\% & 97.7\% \\
Back  window& 99.5\% & 93.9\% & 96.7\% \\
\hline
\end{tabular}
\end{center}
\caption{Accuracy of the sub-segmented parts measured against hand
annotated ground truth.}
\label{tbl:accuracy}
\vskip -5mm
\end{table}
 \paradot{Benchmark tests}
Our results from \emph{Model Assisted Segmentation} were compared with  state of the art image segmentation methods `Grabcut (GC)' \cite{grabcut} and `Level set (LS)' \cite{li2005level} which do not use any \emph{Model Assistance}.
A bounding box has been used initialise the benchmark methods.
We compare our results (Figures \ref{resultSANY2382_800x600w_backdoor_result} and \ref{resultSANY2382_800x600w_fender_result}) with the benchmark tests in Figure \ref{fig:Contours2n3}.
The segmentation using our method are more accurate in general.
In addition to this, our method has the added advantage of sub-segmenting parts of the same object.
This is a non-trivial task for conventional segmentation methods when the sub-segments of the object share the same colour and texture.
In terms of overall performance, we observe that in our method the segmentation results  `bleed' a lot less into adjacent areas, unlike with the benchmark results.
In terms of sub-segmenting parts of the same object, we see in Figure \ref{resultSANY2382_800x600w_fender_result} that our method is capable of successfully segmenting out the fender, mud guard and the buffer from the front door unlike the benchmark methods.
In fact it would be extremely difficult (if not impossible) to sub-segment parts of the front of the car which are painted the same color with conventional methods.
Similarly the back door, back window and the smaller glass panel have been segmented out in Figure \ref{resultSANY2382_800x600w_backdoor_result} where as the benchmark methods group them together.
Results for a semi-profile view of the car are shown in Figure \ref{fig:Contours4} with close ups and benchmark comparisons in  Figure \ref{fig:Contours4n5n6}.
Our results are better and separate the object into meaningful parts.

\section{Discussion}

The \emph{Model Assisted Segmentation} method described in this paper can segment parts of a known 3D object from a given image.
It performs better than the state of the art and can segment (and separate) parts that have similar pixel characteristics.
We present our results on images of cars.
The highly reflective surfaces of cars make the pose estimation as well as the segmentation tasks more difficult than with non-reflective objects.

We note that a close initialisation curve obtained from the 3D pose estimation significantly improves the performance of contour detection, and hence the image segmentation.
However, the presence of reflections can deteriorate the quality of the results.
We intend to explore avenues to make the process more robust in the presence of reflections.

\paradot{Acknowledgment}
The authors wish to thank Stephen Gould and
Hongdong Li for the valuable feedback and advice.
This work was supported by Control\Euro xpert.


\begin{footnotesize}
\bibliographystyle{plain}

\end{footnotesize}

\end{document}